\documentclass[final]{cvpr}

\usepackage{times}
\usepackage{epsfig}
\usepackage{graphicx}
\usepackage{amsmath}
\usepackage{amssymb}
\usepackage{mathrsfs}
\usepackage{multirow}
\usepackage{booktabs}
\usepackage[ruled]{algorithm2e}
\newcommand{\tabincell}[2]{\begin{tabular}{@{}#1@{}}#2\end{tabular}} 
\usepackage[pagebackref=true,breaklinks=true,colorlinks,bookmarks=false]{hyperref}

\def\cvprPaperID{3953} 
\def\confYear{CVPR 2021}

\begin{document}

\title{SwiftNet: Real-time Video Object Segmentation}\vspace{-3mm}

\author{Haochen Wang\textsuperscript{1}\footnotemark[1] , Xiaolong Jiang\textsuperscript{1}\footnotemark[1] , Haibing Ren\textsuperscript{1}, Yao Hu\textsuperscript{1}, Song Bai\textsuperscript{1,2}\\
\textsuperscript{1}Alibaba Youku Cognitive and Intelligent Lab \\
\textsuperscript{2}University of Oxford\\
{\tt\small \{zhinong.whc, xainglu.jxl, haibing.rhb, yaoohu\}@alibaba-inc.com}\\
{\tt\small songbai.site@gmail.com}}

\maketitle
\renewcommand{\thefootnote}{\fnsymbol{footnote}}

\begin{abstract}
In this work we present SwiftNet for real-time semisupervised video object segmentation (one-shot VOS), which reports 77.8\% $\mathcal{J}$\&$\mathcal{F}$ and 70 FPS on DAVIS 2017 validation dataset, leading all present solutions in overall accuracy and speed performance. We achieve this by elaborately compressing spatiotemporal redundancy in matching-based VOS via Pixel-Adaptive Memory (PAM). Temporally, PAM adaptively triggers memory updates on frames where objects display noteworthy inter-frame variations. Spatially, PAM selectively performs memory update and match on dynamic pixels while ignoring the static ones, significantly reducing redundant computations wasted on segmentation-irrelevant pixels. To promote efficient reference encoding, light-aggregation encoder is also introduced in SwiftNet deploying reversed sub-pixel. We hope SwiftNet could set a strong and efficient baseline for real-time VOS and facilitate its application in mobile vision. The source code of SwiftNet can be found at https://github.com/haochenheheda/SwiftNet. 

\end{abstract}
\renewcommand{\thefootnote}{\fnsymbol{footnote}}
\footnotetext[1]{These authors contribute equally.}

\begin{figure}[t]
\begin{center}
\includegraphics[width=1\linewidth]{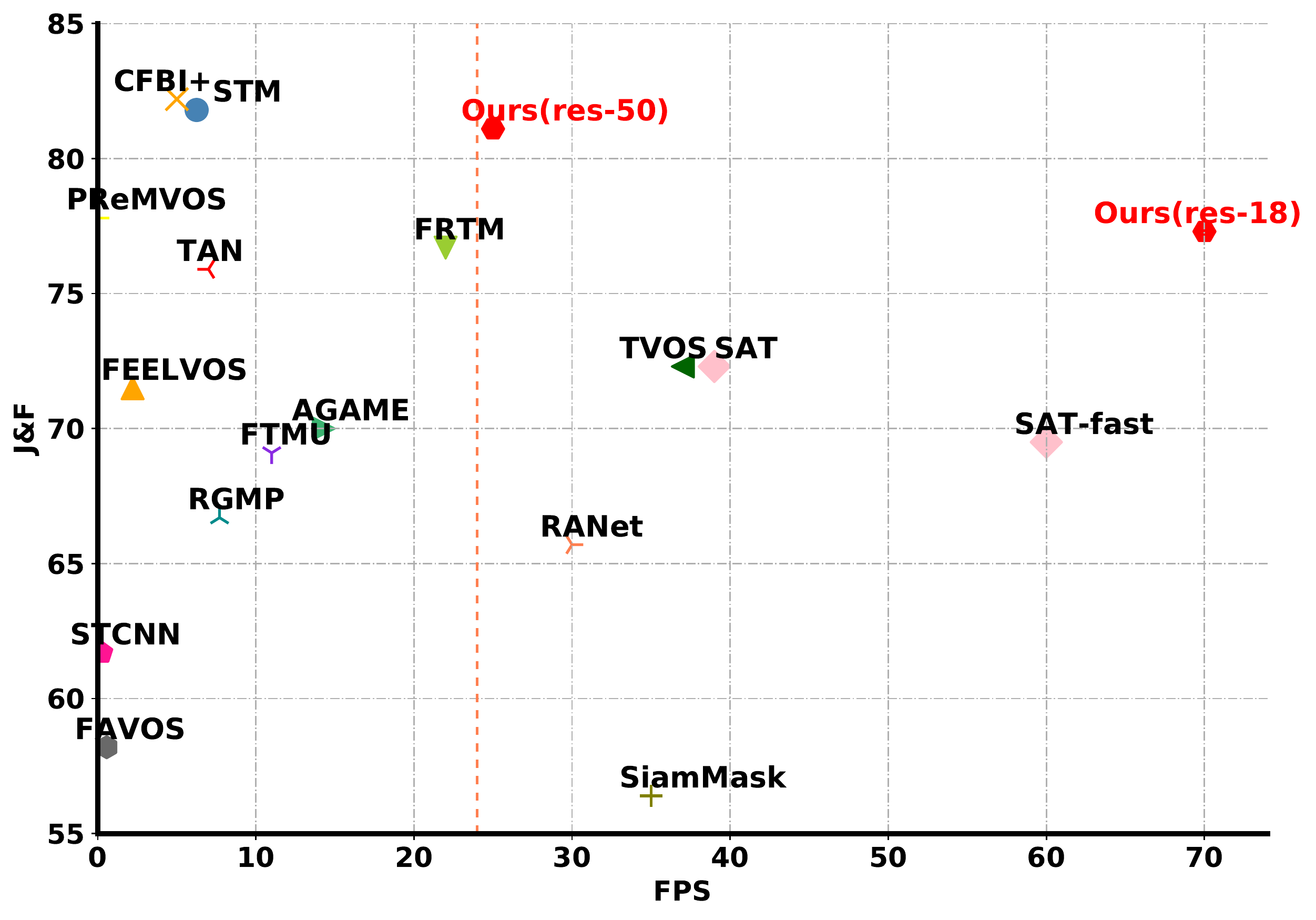}
\end{center}
   \vspace{-5mm}
\caption{Accuracy and speed performance of state-of-the-art methods on DAVIS2017 validation dataset, methods locate on the right side of the red vertical dotted line meet the real-time requirement. Our solutions (ResNet-18 and 50 versions) are marked in red.}
\label{fig:fig1}
\vspace{-2mm}
\end{figure}

\vspace{-3mm}
\section{Introduction}
\vspace{-2mm}
Given the first frame annotation, semi-supervised video object segmentation (one-shot VOS) localizes the annotated object(s) on pixel-level throughout the video. One-shot VOS generally adopts a matching-based strategy, where target objects are first modeled from historical reference frames, then precisely matched against the incoming query frame for localization. Being a video-based task, VOS finds vast applications in surveillance, video editing, and mobile visions, most of which ask for real-time processing \cite{Survey1}. 

Nonetheless, although pursued in fruitful endeavors \cite{OSVOS, Masktrack, RGMP, MotionGuided, Premvos, Feelvos, Agame}, real-time VOS remains unsolved, as object variation over-time poses heavy demands for sophisticated object modeling and matching computations. As a compromise, most existing methods solely focus on improving segmentation accuracy while at the expense of speed. Amongst, memory-based methods \cite{STM, STMr1, STMr2} reveal exceptional accuracy with comprehensively modeling object variations using all historical frames and expressive non-local \cite{NonLocal} reference-query matching. Unfortunately, deploying more reference frames and complicated matching scheme inevitably slow down segmentation. Accordingly, recent attempts seek to accelerate VOS with reduced reference frames and light-weight matching scheme \cite{Fasttan, Fasttmu, GCNet, SAT, RANet, Feelvos, FAVOS, RGMP}. For the first aspect, solutions proposed in \cite{Fasttmu, GCNet, SAT, RANet, Feelvos, FAVOS, RGMP} follow a mask-propagation strategy, where only the first and last historical frames are considered reference for current segmentation. For the second aspect, efficient pixel-wise matching \cite{GCNet, Feelvos, RANet}, region-wise distance measuring \cite{Fasttmu, Fasttan, FAVOS}, and correlation filtering \cite{SAT, SiamMask} are deployed to reduce computations. However, as shown in Fig.\ref{fig:fig1}, although these accelerated methods enjoy faster segmentation speed, they still barely meet real-time requirement, and more critically, they are far from state-of-the-art segmentation accuracy.

\begin{figure}
\begin{center}
\includegraphics[width=1\linewidth]{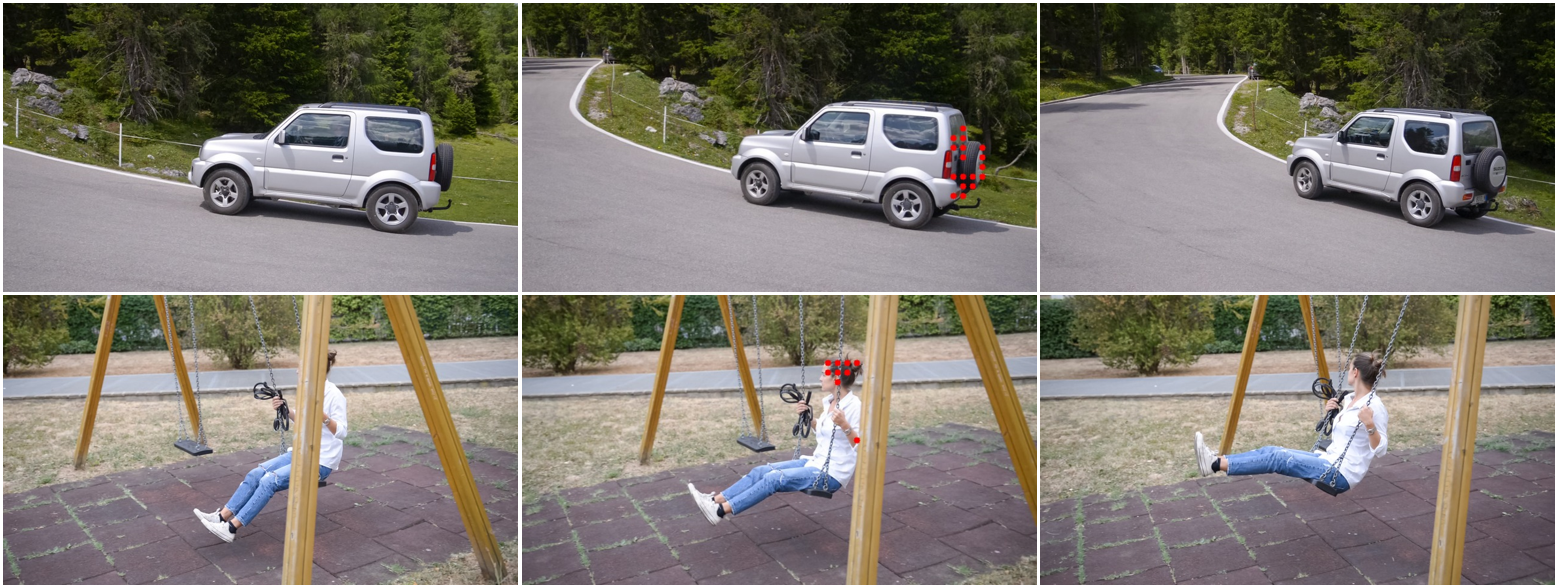}
\end{center}
   \vspace{-3mm}
\caption{An illustration of pixel-wise memory update in video clips. In each row, pixels updated into memory (red) are the ones previously invisible and remain static afterwards.}
\label{fig:fig2}
\vspace{-2mm}
\end{figure}

We argue that, the accurate solutions are less efficient due to the spatiotemporal redundancy inherently resides in matching-based VOS, and the fast solutions suffer degraded accuracy for reducing the redundancy indiscriminately. Considering its pixel-wise modeling, matching, and estimating nature, matching-based VOS manifests positive correlation between processing time $T$ and multiplied number of pixels $\mathcal{N}_r$ and $\mathcal{N}_q$ in reference and query frames as described in Eqa.\ref{equa:equa1}. The spatiotemporal redundancy denotes that $\mathcal{N}_r$ is populated with pixels not beneficial for accurate segmentation. Temporally, existing methods \cite{STM,STMr2} carelessly involve all historical frames (mostly by periodic sampling) for reference modeling, resulting in the fact that static frames showing no object evolution are repeatedly modeled, while dynamic frames containing incremental object information are less attended. Spatially, full-frame modeling and matching are adopted as defaults \cite{STM,cfbi}, wherein most static pixels are redundant for segmentation. Illustration in Fig. \ref{fig:fig2} vividly advocates the above point. From this standpoint, explicitly compressing pixel-wise spatiotemporal redundancy is the best way to yield accurate and fast one-shot VOS.  

\vspace{-3mm}
\begin{equation}
\begin{aligned}
\mathcal{T} \propto \mathcal{O}(\mathcal{N}_{r}\mathop{\times} \mathcal{N}_{q}),
\end{aligned}
\label{equa:equa1}\vspace{-2mm}
\end{equation}

\begin{figure*}
\begin{center}
\includegraphics[width=0.9\linewidth]{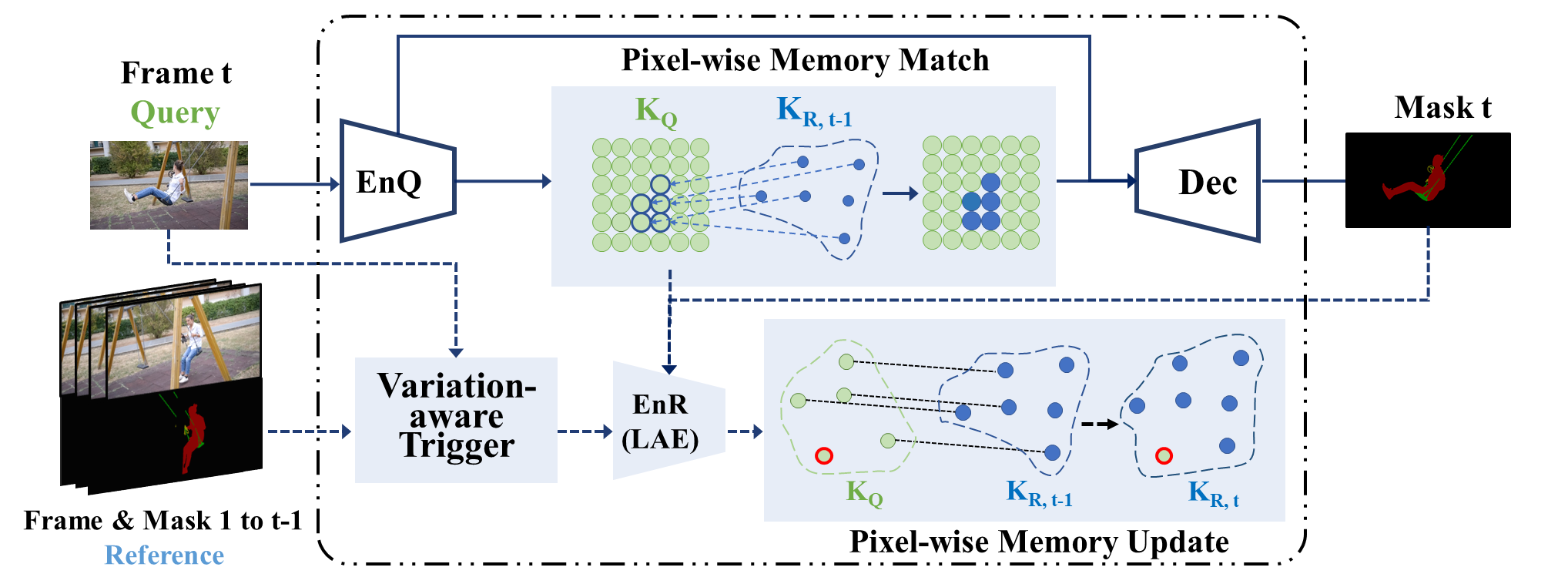}
\end{center}
   \vspace{-3mm}
\caption{An illustration of SwiftNet. Solid black lines represent operation executed first to generate segmentation mask, dotted lines perform afterwards for memory update. Colored dots in memory update and match modules denote pixel-wise key feature vectors. Corresponding value features are omitted in memory update and match for brevity.}
\label{fig:pipe}
\vspace{-2mm}
\end{figure*}

Accordingly, we propose SwiftNet for real-time one-shot video object segmentation. Overall, as depicted in Fig.\ref{fig:pipe}, SwiftNet instantiates matching-based segmentation with an encoder-decoder architecture, where spatiotemporal redundancy is compressed within the proposed Pixel-Adaptive Memory (PAM) component. Temporally, instead of involving all historical frames indiscriminately as reference, PAM introduces a variation-aware trigger module, which computes inter-frame difference to adaptively activate memory update on temporally-varied frames while overlook the static ones. Spatially, we abolish full-frame operations and design pixel-wise update and match modules in PAM. For pixel-wise memory update, we explicitly evaluate inter-frame pixel similarity to identify a subset of pixels beneficial for memory, and incrementally add their feature representation into the memory while bypassing the redundant ones. For pixel-wise memory match, we compress the time-consuming non-local computation to accommodate the pixel-wise memory as reference, thus achieving efficient matching without degradation of accuracy. To further accelerate segmentation, PAM is equipped with a novel light-aggregation encoder (LAE), which eschews redundant feature extraction and enables multi-scale mask-frame aggregation leveraging reversed sub-pixel operations.

In summary, we highlight three main contributions:
\begin{itemize}
\vspace{-2mm}
\item We propose SwiftNet to set the new record in overall segmentation accuracy and speed, thus providing a strong baseline for real-time VOS with publicized source code.
\item We pinpoint spatiotemporal redundancy as the Achilles heel of real-time VOS, and resolve it with Pixel-Adaptive Memory (PAM) composing variation-aware trigger and pixel-wise update \& matching. Light-Aggregation Encoder (LAE) is also introduced for efficient and thorough reference encoding.
\item We conduct extensive experiments deploying various backbones on DAVIS 2016 \& 2017 and YouTube-VOS datasets, reaching the best overall segmentation accuracy and speed performance at 77.8\% $\mathcal{J}$\&$\mathcal{F}$ and 70 FPS on DAVIS2017 validation set.
\end{itemize}

\vspace{-3mm}
\section{Related Work}
\subsection{One-shot VOS}
\vspace{-2mm}
One-shot VOS establishes a spatiotemporal matching problem, such that objects annotated in the first frame are localized in upcoming query frames by searching pixels best-matched to object template modeled in the reference frames. From this perspective, we categorize one-shot VOS methods {\it w.r.t} different reference modeling and reference-query matching strategies. Reference modeling builds object template by exploiting object evolution in historical frames, and methods either follows the last-frame or all-frame approaches. For the former one, \cite{RGMP, RANet, DMMNet, DTN, Agss, Feelvos, Fusionseg} utilize only the first and/or last frame as reference, demonstrate favorable segmentation speed but suffer uncompetitive accuracy due to inadequate modeling over object variation. For the latter one, methods proposed in \cite{STM, STCNN, RVOS, Maskrnn, Vosvm, Tvos, Frtmvos} leverage all previous frames and reveal improved accuracy, but they suffer slower speed for heavy computation overhead even with periodic sampling. 

Considering reference-query matching, we classify methods as two-stage \cite{DTN, MasktrackRcnn, DMMNet} and one-stage \cite{Agss, RGMP, RANet, STCNN, STM, Fusionseg, Feelvos, RVOS, CRN} basing on whether matching with proposed region-of-interest. Similar to object detection, two-stage method is more accurate while one-stage leads in speed. Besides, the key of matching is similarity measuring, where convolutional networks \cite{Agss, RGMP, STCNN, Fusionseg}, cross correlation \cite{RANet, SiamMask}, and non-local computation \cite{STM,  STMr1, STMr2, Feelvos} are widely adopted. Amongst, non-local \cite{NonLocal} reveals best accuracy for capturing all-pairs pixel-wise dependency but are computationally heavy. In addition to matching-based VOS, propagation-based methods \cite{Agame, Masktrack, OSNM, Tvos, pixelwisematch} leverage temporal motion consistency to reinforce segmentation, which is highly effective when appearance matching fails due to severe variations. Additionally, time-consuming online fine-tuning are exploited in \cite{OSVOS,OSVOSpami,anOSVOS} to improve segmentation accuracy, which however is impractical for real-time application.

\subsection{Fast VOS}
For efficiency, most fast VOS solutions deploy the single-frame reference strategy \cite{Feelvos, Agame, RANet, RGMP, OSNM}. Besides, methods proposed in \cite{SiamMask, SAT, FAVOS, Fasttmu, Fasttan} employ segmentation-by-tracking where pixel-wise estimation is gated within tracked bounding-boxes to avoid full-frame estimation. To expedite time-consuming pixel-wise matching, RGMP \cite{RGMP} computes similarity responses with convolutions; AGAME \cite{Agame} discriminates object from background with a probabilistic generative appearance model; RANet \cite{RANet} adopts cross-correlation on ranked pixel-wise features to match query with reference. In addition, OSNM \cite{OSNM} proposes to spur VOS with network modulation.   

\subsection{Memory-based VOS}
Memory-based VOS exploits all historical frames in an external memory for object modeling, an alternative approach for modeling all-frame evolution is via the implementation of recurrent neural networks \cite{Maskrnn, stos, RVOS}. First proposed in \cite{STM}, STM is the seminal memory-based method which boosts segmentation accuracy by a large margin. As follows, \cite{STMr1, STMr2} modify STM by introducing Siamese-based semantic similarity and motion-guided attention. To induce heavy computations, GCNet \cite{GCNet} designs a global context module using attentions to reduce temporal complexity executed in the memory.

\section{SwiftNet}
In this section, we present SwiftNet by first briefly formulating the problem of matching-based one-shot VOS. As follows, PAM is discussed in details, including variation-aware trigger as well as pixel-wise memory update and match modules. LAE is explained afterwards. 

\subsection{Problem Formulation}
Given a video sequence $V = [x_{1}, x_{2}, \cdots, x_{T}]$, its first frame $x_{1}$ is annotated with mask $y_{1}$. The goal of one-shot VOS is to delineate objects from the background by generating mask $y_{t}$ for each frame $t$. Particularly, matching-based VOS computes mask via object modeling and matching. 

For object modeling at frame $t$, historical information embedded in reference frames $[x_{1}, \cdots, x_{t-1}]$ and $[y_{1}, \cdots, y_{t-1}]$ is exploited to establish object model $M_{t-1}$ for up till frame $t-1$:
\begin{equation}
\begin{aligned}
M_{t-1} = \phi(I(1) \cdot EnR(x_{1}, m_{1}), I(2) \cdot EnR(x_{2}, m_{2}), \\
\cdots, I(t-1) \cdot EnR(x_{t-1}, m_{t-1})),
\end{aligned}
\label{equa:equa2}
\end{equation}
here $I(t)$ is an indicator function denoting whether frame $t$ involves in modeling, $EnR(\cdot)$ indicates reference encoder for feature extraction, and $\phi(\cdot)$ generalizes the object modeling process. 

For reference-query matching, the task is to search $M_{t-1}$ within $x_{t}$ on a pixel-level and generate object affinity map $A_{t}$:
\begin{equation}
\begin{aligned}
A_{t} = \gamma(M_{t-1}, EnQ(x_t)),
\end{aligned}
\label{equa:equa3}\vspace{-2mm}
\end{equation}
Here $\gamma(\cdot)$ denotes pixel-wise matching and $EnQ(\cdot)$ refers to the query encoder. The final segmentation mask is produced by a decoder integrating encoded features and $A_{t}$. 

At test time with SwiftNet, upon the arrival of query frame $x_{t}$, it is first processed by the query encoder and then passed into the pixel-wise memory match module. The matching output and encoded query features are aggregated in the decoder to generate mask $y_{t}$. Subsequently, $x_{t}$, $y_{t}$, $x_{t-1}$ and $y_{t-1}$ are jointly fed into the variation-aware trigger module, and if triggered, they are then handled by LAE for later pixel-wise memory update. This overall workflow is illustrated in Fig.\ref{fig:pipe}.  

\subsection{Pixel-Adaptive Memory}
As the core component of SwiftNet, PAM models object evolution and performs object matching with explicitly compressed spatiotemporal redundancy. PAM mainly composes the variation-aware trigger as well as the pixel-wise memory update and match modules. 

\subsubsection{Variation-Aware Trigger}
Instead of utilizing merely the first and last frames for object modeling, incorporating all historical frames as reference help establish temporally-coherent object evolution \cite{STM,stos}. Nonetheless, this approach is rather impractical considering its prohibitive temporal redundancy and computation overhead. As a straightforward solution, previous methods sample historical frames at a predefined pace \cite{STM,Frtmvos}, which indiscriminately reduces temporal redundancy and leads to accuracy degradation. 

To explicitly compress temporal redundancy, variation-aware trigger module evaluates inter-frame variation frame-by-frame, and activates memory update once the accumulated variation surpass threshold $P_{th}$. Specifically, given $x_{t}$, $y_{t}$ and $x_{t-1}$ and $y_{t-1}$, we separately compute image difference $D_f$ and mask difference $D_m$ as:
\vspace{-2mm}
\begin{equation}
\begin{aligned}
D_f^{i} = \sum_{c \in \{R,G,B\} } \mathop{|}x^{i,c}_{t} - x^{i,c}_{t-1}\mathop{|}\mathop{/}255,
\end{aligned}
\label{equa:equa4}\vspace{-2mm}
\end{equation}

\begin{equation}
\begin{aligned}
D_m^{i} = \mathop{|}y^{i}_{t} - y^{i}_{t-1}\mathop{|},
\end{aligned}
\label{equa:equa5}\vspace{-1mm}
\end{equation}
at each pixel $i$ we update the overall running variation degree $P$ as: 
\vspace{-2mm}
\begin{equation}
P =
\begin{cases}
  P + 1, & \text{if}\ D_f^{i} > th_f \ \text{or}\ D_m^{i} > th_m\\
  P, & \text{otherwise}
\end{cases}
\end{equation}
Once $P$ exceeds $P_{th}$, PAM triggers a new round of memory update as described in \ref{sec:bankupdate}. Empirically, $P_{th}$,  $th_f$ and $th_m$ equal 200, 1, 0 respectively yields best performance. 

\begin{figure}
\begin{center}
\includegraphics[width=1\linewidth]{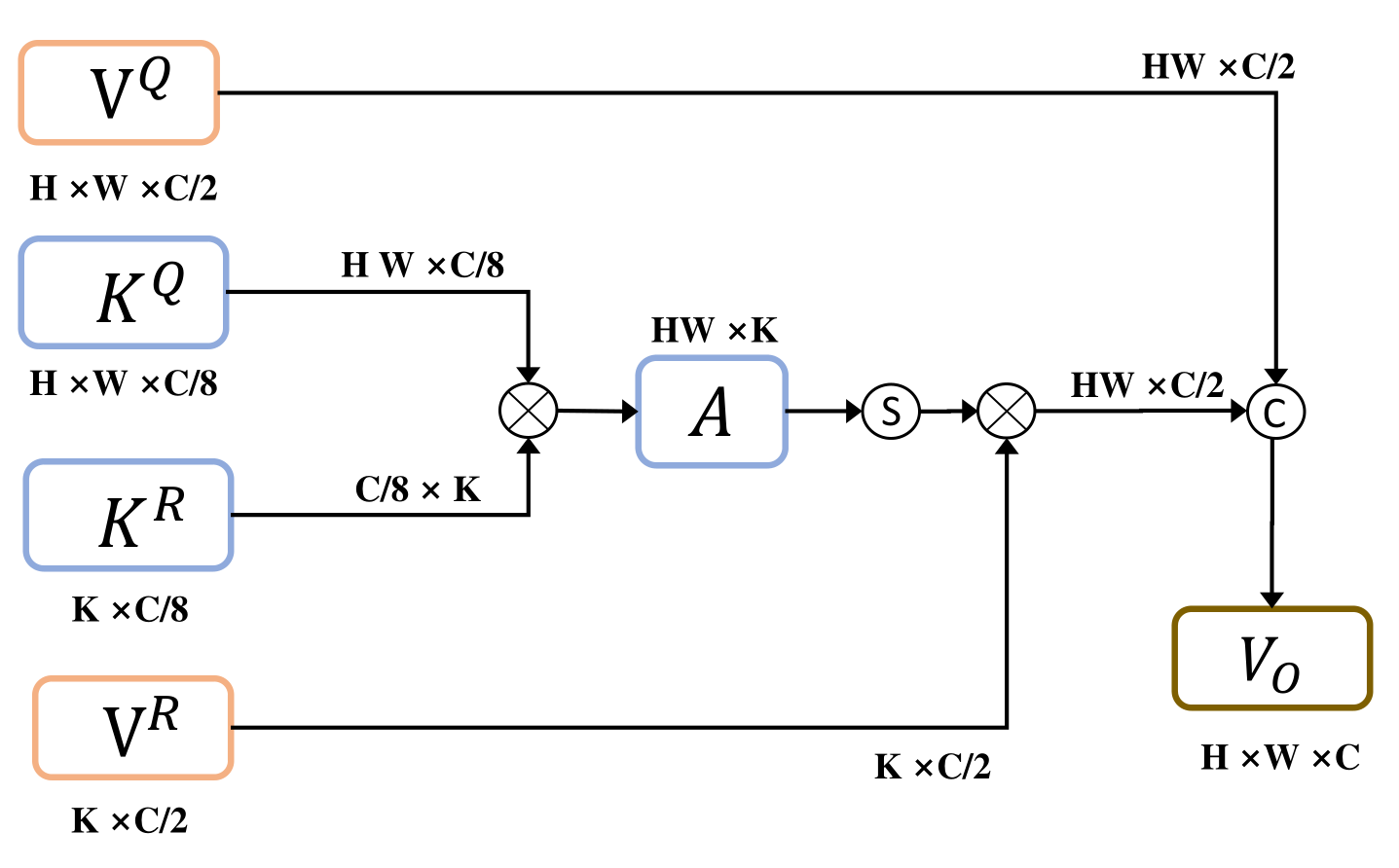}
\end{center}
   \vspace{-5mm}
\caption{A diagram of the pixel-wise memory match computation, sub-script t is omitted for brevity. The dotted s and c stands for Softmax and concatenation operations.}
\label{fig:NL}
\vspace{-2mm}
\end{figure}

\subsubsection{Pixel-wise Memory Update}
\label{sec:bankupdate}
In terms of matching-based VOS, memory infers a temporally-maintained template which characterizes object evolution over time. In the existing literature, memory update and matching typically adopt full-frame operations, where reference frames are concatenated into memory and matched with query frame intactly \cite{STM,STMr2}. This strategy induces heavy storage and computation overhead, as redundant pixels showing no benefit for object modeling are incorporated without discrimination. 

To compress redundant pixels from full frames, PAM introduces pixel-wise memory update and match modules. For memory update, if frame $x_{t}$ is triggered, PAM first discovers pixels in $x_{t}$ that demonstrates significant variations from itself in memory $M_{t-1}$, then incrementally updates newly discovered features (as displayed in $x_{t}$) into the memory. Through $EnR$, $x_{t}$ is encoded into key $K_{Q,t} \in\mathbb{R}^{H \times W \times C/8}$ and value $V_{Q,t} \in\mathbb{R}^{H \times W \times C/2}$ features, key features are with shallower depth to facilitate efficient matching. In the experiment $C$ is set to 256. Similarly, memory $M_{t-1}$ containing $k_{t-1}$ pixels in forms of $K_{R,t-1} \in\mathbb{R}^{k_{t-1} \times C/8}$ and $V_{R,t-1} \in\mathbb{R}^{k_{t-1} \times C/2}$. To discover varied pixels, we flatten $K_{Q,t}$ and compute cosine similarity matrix $S \in\mathbb{R}^{HW \times k_{t-1}}$ as:
\vspace{-2mm}
\begin{equation}
\begin{aligned}
S^{i,j} = \frac{K^{i}_{Q,t} \cdot K^{j}_{R,t-1}}{\|K^{i}_{Q,t}\| \|K^{j}_{R,t-1}\|},
\end{aligned}
\label{equa:sim1}\vspace{-2mm}
\end{equation}
for each row $i$ in $S$, we find the largest score as the feature similarity between pixel $i$ in query frame $t$ and in pixel $j$ in the memory. In formulation, we compute pixel similarity vector $s$ as:
\begin{equation}
\begin{aligned}
s_i = \mathop{\arg\max}_{j}S[i,:],
\end{aligned}
\label{equa:vec}\vspace{-2mm}
\end{equation}
we sort $s$ in increasing order of similarity (original index is kept), then the select top $\beta$ percents pixels for memory update. These set of pixels exhibit most severe feature variations. Here $\beta$ is a hyper-parameter controlling the balance between method efficiency and update comprehensiveness, and is experimentally set to 10\% for the best performance. To execute the memory update, we find feature vectors of the selected set of pixels from $K_{Q,t}$ and $V_{Q,t}$ according to indexes as in $s$, then directly add them into memory $M$ which is instantiated as an array of feature vectors. 

\begin{figure}
\begin{center}
\includegraphics[width=1\linewidth]{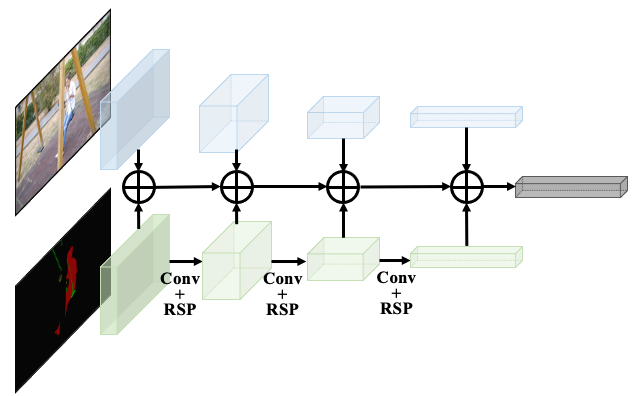}
\end{center}
\caption{An illustration of LAE. Image features are pre-computed by $EnQ$, mask features are parallelly generated via convolutions and revered sub-pixel (RSP) modules.}
\label{fig:encoder}
\vspace{-2mm}
\end{figure}

\subsubsection{Pixel-wise Memory Match}
As illustrated in Fig. \ref{fig:pipe}, segmentation mask is decoded utilizing query value $V_{Q}$ and output of reference-query matching, which provides strong spatial prior {\it w.r.t.} the foreground object. In essence, this matching process computes similarity between pixels from the reference and query frames, and can be instantiated with cross-correlation \cite{SAT,SiamMask}, neural networks \cite{RGMP,Agame}, distance measuring \cite{Feelvos}, and non-local computation \cite{STM}, etc. Comparatively, non-local leads to excellent accuracy performance but suffers heavy computation expenses in the context of full-frame operations. In PAM, we implement pixel-wise matching to achieve efficient and accurate segmentation. 

In Fig. \ref{fig:NL} we illustrate the pipeline of pixel-wise matching computation. At first, query frame key $K_{Q,t}$ and the memory key $K_{R,t-1}$ are reshaped into vectors of size $HW \times C/8$ and $C/8 \times K$. We then calculate dot-product similarity between corresponding vectors to produce affinity map $A \in\mathbb{R}^{HW \times K}$ as: 
\begin{equation}
\begin{aligned}
A^{i,j} = \text{exp}(K^{i}_{Q,t} \odot K^{j}_{R,t-1}),
\end{aligned}
\label{equa:sim2}\vspace{-2mm}
\end{equation}
$I_{t}$ is passed through a Softmax layer and further multiplied with memory value $V_{R,t-1}$. The resulted tensor ($\in\mathbb{R}^{HW \times C/2}$) is concatenated with $V_{Q,t}$ to form the activated feature as:
\begin{equation}
\begin{aligned}
V_D = \mathop{[}A \mathop{\times} V_{R,t-1},V_{Q,t}\mathop{]},
\end{aligned}
\label{equa:sim2}\vspace{-2mm}
\end{equation}
which is then input into the decoder. We emphasize that, our approach is different to normal non-local computation implemented in \cite{STM} as we eliminate redundant pixels from full-frame memory. Affinity map $A$, being the computation bottleneck with size $\mathbb{R}^{HW \times HWT}$, is significantly reduced to $\mathbb{R}^{HW \times K}$. $K$ is the size of pixel-wise memory and is strongly controlled by the update ratio $\beta$. By explicitly compressing redundancy during memory update and matching, storage requirement and computation speed are both optimized in SwiftNet without considerable accuracy loss.

\subsection{Light-Aggregation Encoder}
In existing memory-based VOS solutions \cite{STM,STMr2}, the image encoding process is time-consuming as both query and reference encoders adopt heavy backbone networks. In SwiftNet we expedite encoding by sparing the backbone resides in $EnR$. Particularly, we instantiate $EnQ$ with ResNet-based \cite{Resnet} networks, and after $x_t$ is encoded by $EnQ$, we buffer the generated feature maps. If frame $t$ is triggered for update, these buffered features are directly utilized in $EnR$ for reference encoding. Efficiency comparison {\it w.r.t.} encoding strategies are listed in Table. \ref{tab:encodercheck}. 


To facilitate the described encoding process, we design the novel light-aggregation encoder as shown in Fig. \ref{fig:encoder}. The upper blue entities represent buffered feature maps encoded by $EnQ$, the bottom green ones show feature transformation hierarchy of the input mask. Features aligned vertically in the same column are with the same size and concatenated together to facilitate multi-scale aggregation. In particular, to instantiate feature transformation of the input mask, we implement reversed sub-pixel for down-samplings and $1 \times 1$ convolutions for channel manipulation. Reversed sub-pixel technique is motivated by the popular up-sampling method in super-resolution \cite{subpixel}, which shrinks spatial dimension of features without information loss.

\section{Experiments}
In this section we first discuss implementation details of the experiments, then elaborate the ablation study specifying contributions of different components proposed in SwiftNet. Comparisons with other state-of-the-art methods on DAVIS 2016 \& 2017 and YouTube-VOS datasets are provided as follows, where SwiftNet demonstrates the best overall segmentation accuracy and inference speed. All experiments are implemented in PyTorch \cite{pytorch} on 1 NVIDIA P100 GPU. Particularly, SwiftNet adopting both ResNet-18 and ResNet-50 \cite{Resnet} backbones are experimented to show the favorable compatibility and efficacy of our method. We employs a decoder similarly constructed as in STM ~\cite{STM}, which adopts three refinement modules to gradually restore the spatial scale of the segmentation mask.

\subsection{Datasets and Evaluation Metrics}
\noindent\textbf{DAVIS 2016 \& 2017}. DAVIS 2016 dataset contains in total 50 single-object videos with 3455 annotated frames. Considering its confined size and generalizability, it is soon supplemented into DAVIS 2017 dataset comprising 150 sequences with 10459 annotated frames, a subset of which exhibit multiple objects. Following the DAVIS standard, we utilize mean Jaccard $\mathcal{J}$ index and mean boundary $\mathcal{F}$ score, along with mean $\mathcal{J}$\&$\mathcal{F}$ to evaluate segmentation accuracy. We adopt the Frames-Per-Second (FPS) metric to measure segmentation speed.

\begin{table}[h]
\small
\centering
\begin{tabular}{l c c c c }
\toprule
\multirow{2}*{Method} & \multicolumn{2}{c}{w/o pixel-wise} & \multicolumn{2}{c}{w pixel-wise}\\
\cmidrule(lr){2-3} \cmidrule(lr){4-5}
&J\&F&FPS&J\&F&FPS\quad\\\midrule
low-level&78.0&22&77.5&51\\
high-level&75.4&37&73.6&71\\
LAE&78.2&35&77.8&70\\\bottomrule
\end{tabular}
\vspace{2mm}
\caption{Ablation study of LAE on Davis 2017 validation set.}
\label{tab:encodercheck}
\end{table}

\noindent\textbf{YouTube-VOS}. Being the largest dataset at the present, YouTube-VOS encompasses totally 4453 videos annotated with multiple objects. In particular, its validation set possesses 474 sequences covering 91 object classes, 26 of which are not visible in the training set, and thus facilitating evaluations {\it w.r.t.} seen and unseen object classes to reflect method generalizability. On YouTube-VOS we report $\mathcal{J}$\&$\mathcal{F}$ for accuracy assessment, the overall score $\mathcal{G}$ is generated by averaging $\mathcal{J}$\&$\mathcal{F}$ on seen and unseen classes. 

\subsection{Training and Inference}
\subsubsection{Training}
SwiftNet is first pre-trained on simulated data generated upon MS-COCO dataset ~\cite{coco}, then finetuned on DAVIS 2017 and YouTube-VOS Dataset respectively. In both training stages, input image size is set to 384 $\times$ 384, and we adopt Adam optimizer with learning rate starting at 1e-5. The learning rate is adjusted with polynomial scheduling using the power of 0.9. All batch normalization layers in the backbone are fixed at its ImageNet pre-trained value during training. We use batch size of 4, which is realized on 1 GPU via manual accumulation. 

\noindent\textbf{MS-COCO Pre-train}. Considering the scarcity of video data and to ensure the generalizability of SwiftNet, we perform pre-training on simulated video clips generated upon MS-COCO dataset \cite{mscoco}. Specifically, we randomly crop foreground objects from a static image, which are then pasted onto a randomly sampled background image to form a simulated image. Affine transformations such as rotation, resizing, sheering, and translation are applied to foreground and background separately to generate deformation and occlusion, and we maintain an implicit motion model to generate clips with length of 5. SwiftNet is trained with simulated clips for 150000 iterations and the $\mathcal{J}$\&$\mathcal{F}$ reaches 65.6 on DAVIS 2017 validation set, which demonstrates the efficacy of our simulated pre-training.

\noindent\textbf{DAVIS 2017 $\&$ YouTube-VOS Finetune}. After pre-training, we finetune SwiftNet on DAVIS 2017 and YouTube-VOS training set for 200000 iterations. At each iteration, we randomly sampled 5 images consecutively (with random skipping step smaller than 5 frames) and estimate corresponding segmentation masks one after another. Pixel-wise memory update and match are executed on every frame within the 5-frame clip.

\begin{table}[h]
\small
\centering
\begin{tabular}{c c c c}
\toprule
\quad&Metric& \tabincell{c}{periodical \\ sampling (5)} & \tabincell{c}{variation-aware\\ trigger}\\\midrule
\multirow{2}*{full frame}&J\&F&78.2&78.1\\
&FPS&35&52\\\midrule
\multirow{2}*{\tabincell{c}{pixel-wise\\ update \& match}}&J\&F&77.8&77.8\\
&FPS&65&70\\\bottomrule
\end{tabular}
\vspace{2mm}
\caption{Ablation study of PAM on Davis 2017 validation set.}
\label{tab:pixel-wise}
\end{table}

\vspace{-2mm}
\subsubsection{Inference}
Given a test video accompanied by its first frame annotation mask, at inference time we frame-by-frame segment the video using SwiftNet. Particularly, memory at the first frame, $M_0$, is initialized with feature maps output by the encoder given first frame image and mask, then it is updated online throughout the inference. At frame $t$, we utilize memory $M_{t-1}$ and frame image $I_t$ to compute segmentation mask $m_t$ with SwiftNet. If frame $t$ is triggerd, $m_t$ is feed into the LAE and to update the memory for further computations. 

\subsection{Ablation Study}
Ablation study is conducted on DAVIS 2017 validation set to show contributions of different SwiftNet modules.

\begin{figure*}[t]
\begin{center}
\includegraphics[width=1\linewidth]{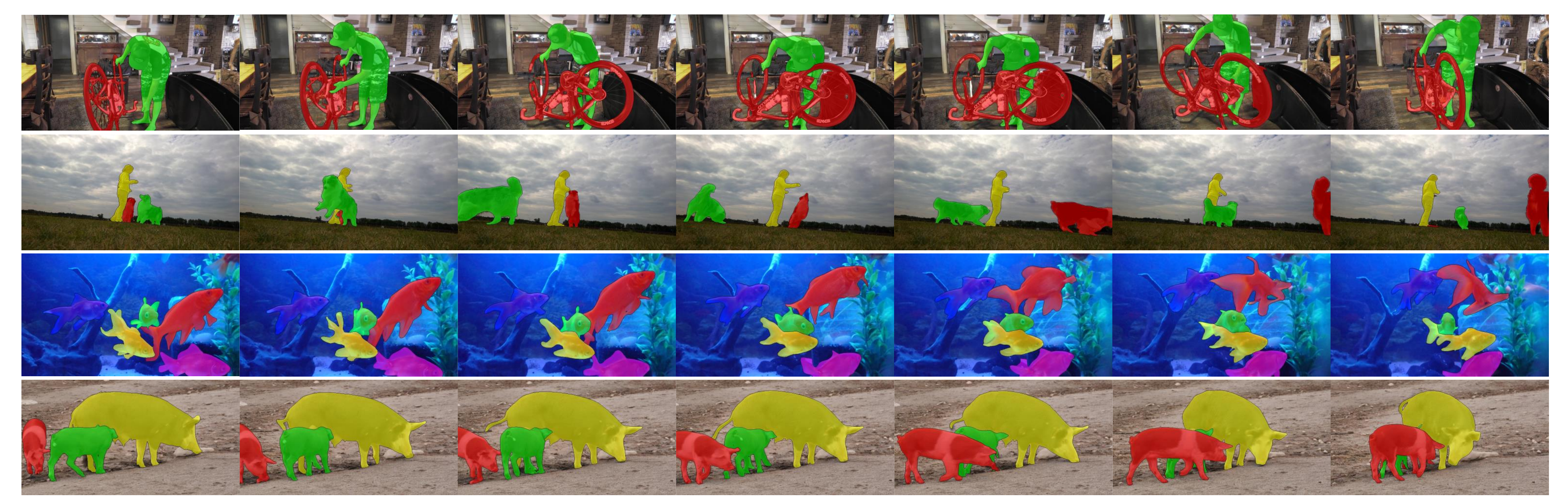}
\end{center}
\caption{Qualitative results of SwiftNet (ResNet-18) generated on DAVIS17 validation set.}
\label{fig:vis}
\vspace{-2mm}
\end{figure*}

\begin{figure}[h]
\begin{center}
\includegraphics[width=0.9\linewidth]{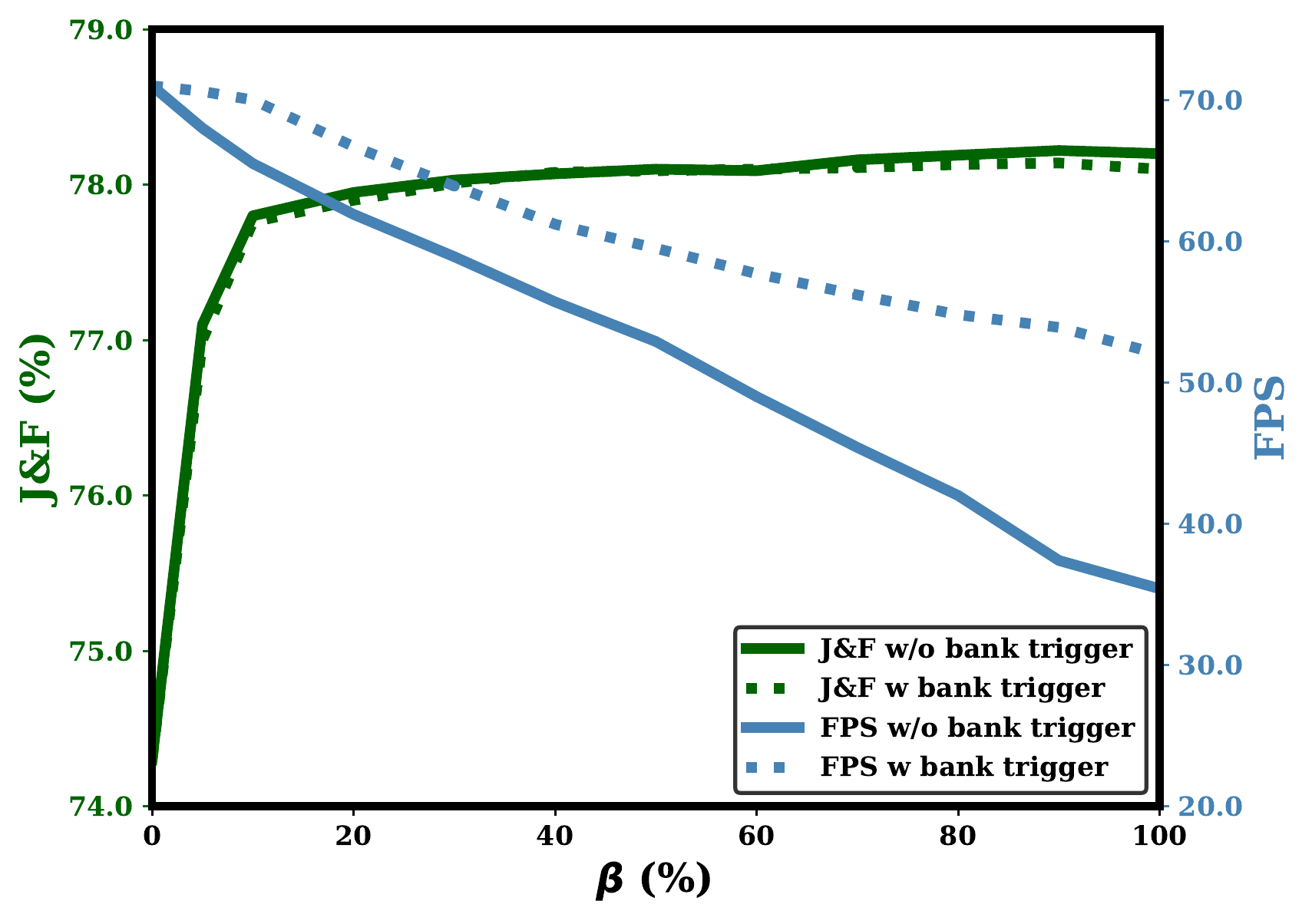}
\end{center}
   \vspace{-4mm}
\caption{$\mathcal{J}$\&$\mathcal{F}$ and FPS evolution {\it w.r.t.} update ratio $\beta$.}
\label{fig:ratio_of_pixels}
\vspace{-4mm}
\end{figure}

\subsubsection{Light-Aggregation Encoder}
To demonstrate the efficacy of the proposed LAE, we additionally develop two baseline reference encoders for comparison. The first baseline instantiates low-level aggregation as adopted in STM ~\cite{STM}, where mask produced by the last frame is directly concatenated with raw image. This encoder maintains high-resolution mask but requires two separate encoders for reference and query frames, hence heavier model size. The second baseline implements high-level aggregation following CFBI ~\cite{cfbi}, where segmentation mask is first down-sampled to the minimal feature resolution and then fused with query feature for foreground discovery. This baseline enables encoder reuse between reference query frames, but spatial details of the mask are lost during pooling-based down-samplings. As shown in Table ~\ref{tab:encodercheck}, low-level baseline reveals better accuracy while high-level baseline runs faster, conforming to the fact that the low-level one involves more sophisticated feature aggregations between image and mask. Notably, LAE surpasses the low-level baseline in both $\mathcal{J}$\&$\mathcal{F}$ and FPS (by 19), and outperforms the high-level baseline by 4.2\% in $\mathcal{J}$\&$\mathcal{F}$ while keeping comparable FPS. This results strongly suggest that LAE promotes thorough mask-frame aggregation and elevates segmentation speed. 

\subsubsection{Pixel-Adaptive Memory}
\vspace{-2mm}
In this section we showcase the efficacy of PAM in elevating accuracy and speed. Table ~\ref{tab:pixel-wise} row-wise illustrates the contribution of pixel-wise memory update and match in eliminating spatial redundancy, where it significantly boosts processing speed by 30 and 18 PFS in both temporal strategies, and only experience around 0.4\% drop in $\mathcal{J}$\&$\mathcal{F}$. Column-wise reveals the contribution of variation-aware trigger in compressing temporal redundancy, where it raises segmentation speed by 17 and 5 FPS in both spatial strategies, and at most 0.1\% $\mathcal{J}$\&$\mathcal{F}$ is reported. Notably here we experiment with periodic sampling at a pace of 5 frame, which is tested to be the optimal parameter as in \cite{STM}. To provide a up-closer view of PAM, in Fig. ~\ref{fig:ratio_of_pixels} we illustrate the variation of $\mathcal{J}$\&$\mathcal{F}$ and FPS {\it w.r.t.} different spatial update ratio $\beta$ and temporal trigger strategy. As shown in green, $\mathcal{J}$\&$\mathcal{F}$ increases in accordance with enlarged $\beta$, {\it i.e.} segmentation accuracy will grow if more percentage of pixels are updated. It is worth noting that, $\beta = 10\%$ yields the best accuracy while larger value shows no significant improvement. Besides, temporal trigger brings minute effect in accuracy. The blue color draws variations {\it w.r.t.} FPS, where larger $\beta$ steadily decreases FPS, and variation-aware trigger constantly increases FPS in under different $\beta$. Notably, the gap between blue curves are enlarged with larger $\beta$, echos that more spatiotemporal redundancy are compressed by the trigger during heavy spatial update. 

\subsection{State-of-the-art Comparison}
\subsubsection{DAVIS 2017}
\vspace{-2mm}
Comparison results on Davis 2017 validation set are listed in Table ~\ref{tab:davis17}. As shown, both SwiftNet versions demonstrate better $\mathcal{J}$\&$\mathcal{F}$, $\mathcal{J}$, and $\mathcal{F}$ scores than all other real-time methods by a large margin. In particular, SwiftNet with ResNet-18 runs the fastest at 70 FPS, outperforming the second fastest SAT-fast \cite{SAT} in $\mathcal{J}$\&$\mathcal{F}$ by 8.3\%. This considerable lead is because that SAT updates global feature with cropped regions containing heavy background noise, while SwiftNet updates memory with useful and discriminative pixels and filters out redundant and noise regions. SwiftNet with ResNet-50 not only meets real-time requirement, but also reaches 81.1 in $\mathcal{J}$\&$\mathcal{F}$ score, which ranks the second best in both real-time and slow methods. STM \cite{STM} reports the best $\mathcal{J}$\&$\mathcal{F}$ at 81.8, which is 0.7\% better than ours, while we run almost 4 times faster than STM. This significant improvement of SwiftNet is achieved by explicitly compressing spatiotemporal redundancy resides in STM, which adopts heavy periodical sampling and full-frame matching. In addition, GCNet \cite{GCNet} also strives to accelerate memory-based VOS by designing light-weight memory reading and writing strategies. As shown, it runs at comparable speed with our ResNet-50 version, while we exceeds GCNet in term of $\mathcal{J}$\&$\mathcal{F}$ by 9.7\%. Fig ~\ref{fig:vis} shows qualitative results on DAVIS17 validation set produced by SwiftNet with ResNet-18. The first row demonstrates that SwiftNet is robust against deformation, the second to the fourth row reveal that SwiftNet is highly capable of handling fast motion, similar distractor, and tremendous occlusion, respectively.

\begin{table}[h]
\small
\centering
\begin{tabular}{l c c c c c}
\toprule
Method & OL & {$\mathcal{J}$\&$\mathcal{F}$} & $\mathcal{J}$ & $\mathcal{F}$ & FPS\\ \midrule
PReMVOS~\cite{Premvos} & $\surd$ & 77.8& 73.9& 81.7 & 0.01\\
CINM~\cite{cinm}&$\surd$ & 67.5&64.5&70.5&0.01\\
OnAVOS~\cite{anOSVOS} & $\surd$ & 67.9&64.5&70.5&0.08\\
OSVOS~\cite{OSVOS}&$\surd$&60.3&56.7&63.9&0.22\\
OSVOS-s\cite{osvoss} & $\surd$ & 68.0&64.7&71.3&0.22\\
STCNN~\cite{STCNN}&$\times$&61.7&58.7&64.6&0.25\\
FAVOS~\cite{FAVOS}&$\times$&58.2&54.6&61.8&0.56\\
FEELVOS~\cite{Feelvos}&$\times$&71.5&69.1&74.0&2.2\\
Dyenet~\cite{dyenet}&$\surd$&69.1&67.3&71.0&2.4\\
STM~\cite{STM}&$\times$&\textbf{81.8}&\textbf{79.2}&\textbf{84.3}&6.3\\
Fasttan~\cite{Fasttan}&$\times$&75.9&72.3&79.4&7\\
RGMP~\cite{RGMP}&$\times$&66.7&64.8&68.6&7.7\\
Fasttmu~\cite{Fasttmu}&$\times$&70.6&69.1&72.1&11\\
AGAME~\cite{Agame}&$\times$&70.0&67.2&72.7&14\\
FRTM-VOS~\cite{Frtmvos}&$\surd$&76.7&-&-&22\\
\midrule
GCNet~\cite{GCNet}&$\times$&71.4&69.3&73.5&25\\
RANet~\cite{RANet}&$\times$&65.7&63.2&68.2&30\\
SiamMask~\cite{SiamMask}&$\times$&56.4&64.3&58.5&35\\
TVOS~\cite{Tvos}&$\times$&72.3&69.9&74.7&37\\
SAT~\cite{SAT}&$\times$&72.3&68.6&76.0&39\\
FRTM-VOS-fast~\cite{Frtmvos}&$\surd$&70.2&-&-&41\\
SAT-fast~\cite{SAT}&$\times$&69.5&65.4&73.6&60\\
\hline\hline
\textbf{SwiftNet(ResNet-50)}&$\times$&\textbf{81.1}&\textbf{78.3}&\textbf{83.9}&25\\
\textbf{SwiftNet(ResNet-18)}&$\times$&77.8&75.7&79.9&\textbf{70}\\\bottomrule
\end{tabular}
\vspace{2mm}
\caption{Quantitative results on DAVIS 2017 validation set. In all following tables, OL denotes online learning and real-time methods reside below the horizontal line.\vspace{-4mm}}
\label{tab:davis17}
\end{table}
\vspace{-3mm}
\subsubsection{DAVIS 2016}\vspace{-1mm}
Results on DAVIS 2016 dataset is shown in Table. \ref{tab:davis16}. Since DAVIS 2016 only contains single-object sequences, most methods experience considerable performance gains when transferred from DAVIS 2017, and the accuracy gap between ResNet-18 and ResNet-50 SwiftNet is reduced because the demands for highly semantical features are alleviated. It is worth noting that, SwiftNet with both ResNet-18 and ResNet-50 outperform all other methods in segmentation accuracy, where the ResNet-50 version leads the second best STM by 1.1\% and 18.7 in terms of $\mathcal{J}$\&$\mathcal{F}$ and FPS. 
\vspace{-6mm}
\subsubsection{Youtube-VOS}\vspace{-2mm}
As testing on the large YouTube-VOS validation set is time-consuming, here we show comparison results with most representative methods. Besides, considering the varied formulation of test sequences, we omit FPS readings on YouTube-VOS by default. As shown in Table \ref{tab:youtube}, SwiftNet with ResNet-50 considerably outperform all other real-time methods in accuracy, leading the second best GCNet by 4.6\% in term of overall score $G$, while running at a comparable speed with GCNet. SwiftNet with ResNet-18 performs comparably with GCNet, but runs significantly faster. Moreover, SwiftNet performs stably across seen and unseen classes, demonstrating its favorable generalizability. 

\begin{table}[tb]
\footnotesize
\centering
\begin{tabular}{l c c c c c}
\toprule
Method & OL & {$\mathcal{J}$\&$\mathcal{F}$} & $\mathcal{J}$ & $\mathcal{F}$ & FPS\\ \midrule
PReMVOS~\cite{Premvos} & $\surd$ & 86.8& 84.9&88.6  & 0.01\\ 
OnAVOS~\cite{onavos} & $\surd$ & 85.5&86.1&84.9&0.08\\
OSVOS~\cite{OSVOS} & $\surd$ & 80.2&79.8&80.6&0.22\\
RANet+~\cite{RANet}& $\surd$ & 87.1 & 86.6 & 87.6 & 0.25\\
FAVOS~\cite{FAVOS}& $\times$ &80.8 & 82.4& 79.5&0.56\\
FEELVOS~\cite{Feelvos}& $\times$ &81.7 & 81.1&82.2 &2.2\\
Dyenet~\cite{dyenet}&$\surd$&-&86.2&-&2.4\\
STM~\cite{STM}& $\times$ & \textbf{89.3} & \textbf{88.7}& \textbf{89.9}&6.3\\
Fasttan~\cite{Fasttan}&$\times$&75.9&72.3&79.4&7\\
RGMP~\cite{RGMP}& $\times$ & 81.8& 81.5& 82.0&7.7\\
Fasttmu~\cite{Fasttmu}&$\times$&78.9&77.5&80.3&11\\
AGAME~\cite{Agame}& $\times$ &- &82.0 &- &14\\
FRTM-VOS~\cite{Frtmvos}&$\surd$&83.5&-&-&22\\
\midrule
GCNet~\cite{GCNet}&$\times$&86.6&87.6&85.7&25\\
SiamMask~\cite{SiamMask}&$\times$&70.0&71.7&67.8&35\\
SAT~\cite{SAT}&$\times$&83.1&82.6&83.6&39\\
FRTM-VOS-fast~\cite{Frtmvos}&$\surd$&78.5&-&-&41\\
\hline\hline
\textbf{SwiftNet(ResNet-50)}&$\times$&\textbf{90.4}&\textbf{90.5}&\textbf{90.3}&25\\
\textbf{SwiftNet(ResNet-18)}&$\times$&90.1&90.3&89.9&\textbf{70}\\\bottomrule
\end{tabular}
\vspace{2mm}
\caption{Quantitative results on DAVIS 2016 validation set.}
\label{tab:davis16}
\end{table}

\begin{table}
\footnotesize
\centering
\begin{tabular}{l c c c c c c}
\toprule
Method & OL & G & $\mathcal{J}_{s}$ & $\mathcal{J}_{u}$ & $\mathcal{F}_{s}$ & $\mathcal{F}_{u}$\\ \midrule
RGMP~\cite{RGMP}&$\times$&53.8&59.5&45.2&-&-\\
OnAVOS~\cite{onavos}&$\surd$&55.2&60.1&46.1&62.7&51.4\\
PReMVOS~\cite{Premvos}&$\surd$&66.9&71.4&56.5&75.9&63.7\\
OSVOS~\cite{OSVOS}&$\surd$&58.8&59.8&54.2&60.5&60.7\\
FRTM-VOS~\cite{Frtmvos}&$\surd$&72.1&72.3&65.9&\textbf{76.2}&74.1\\
STM~\cite{STM}&$\times$&\textbf{79.4}&\textbf{79.7}&\textbf{84.2}&72.8&\textbf{80.9}\\
\midrule
SiamMask~\cite{SiamMask}&$\times$&52.8&60.2&45.1&58.2&47.7\\
SAT~\cite{SAT}&$\times$&63.6&67.1&55.3&70.2&61.7\\
FRTM-VOS-fast~\cite{Frtmvos}&$\surd$&65.7&68.6&58.4&71.3&64.5\\
TVOS~\cite{Tvos}&$\times$&67.8&67.1&63.0&69.4&71.6\\
GCNet~\cite{GCNet}&$\times$&73.2&72.6&68.9&75.6&75.7\\
\hline\hline
\textbf{SwiftNet(ResNet-50)}&$\times$&\textbf{77.8}&\textbf{77.8}&\textbf{72.3}&\textbf{81.8}&\textbf{79.5}\\
\textbf{SwiftNet(ResNet-18)}&$\times$&73.2&73.3&68.1&76.3&75.0\\\bottomrule
\end{tabular}
\vspace{1mm}
\caption{Quantitative results on Youtube-VOS validation set. G denotes overall score, subscript s and u denote scores in seen and unseen categories.\vspace{-2mm}}
\label{tab:youtube}
\end{table}
\vspace{-2mm}

\section{Conclusion}
\vspace{-2mm}
We propose a real-time semi-supervised video object segmentation (VOS) solution, named SwiftNet, which delivers the best overall accuracy and speed performance. SwiftNet achieves real-time segmentation by explicitly compressing spatiotemporal redundancy via Pixel-Adaptive Memory (PAM). In PAM, temporal redundancy is reduced using variation-aware trigger, which adaptively selects incremental frames for memory update and ignores static ones. Spatial redundancy is eliminated with pixel-wise memory update and match modules, which abandon full-frame operations and incrementally process with temporally-varied pixels. Light-aggregation encoder is also introduced to promote thorough and expedite reference encoding. Overall, SwiftNet is effective and compatible, we hope it could set a strong baseline for real-time VOS. 

{\small
\bibliographystyle{ieee_fullname}
\bibliography{cvpr_fastvos_v2.bbl}
}


\end{document}